\documentclass[runningheads]{llncs}
\usepackage[T1]{fontenc}
\usepackage{amsmath}
\usepackage{graphicx}
\usepackage{float}
\usepackage{multirow} 
\usepackage{caption}
\usepackage{wrapfig}
\usepackage{tabularx}
\begin{document}
\title{CycleIK: Neuro-inspired Inverse Kinematics}
%
%\titlerunning{Abbreviated paper title}
% If the paper title is too long for the running head, you can set
% an abbreviated paper title here
%
%\author{First Author\inst{1}\orcidID{0000-1111-2222-3333} \and
%Second Author\inst{2,3}\orcidID{1111-2222-3333-4444} \and
%Third Author\inst{3}\orcidID{2222--3333-4444-5555}}
\author{Jan-Gerrit Habekost\and
Erik Strahl\and
Philipp Allgeuer\and
Matthias Kerzel\and
Stefan Wermter}
\authorrunning{Habekost et al.}
% First names are abbreviated in the running head.
% If there are more than two authors, 'et al.' is used.
%
%\institute{Knowledge Technology, Department of Informatics, University of Hamburg,\\{[ Hamburg, {[ Germany ]}]} \\
\institute{Knowledge Technology, Department of Informatics, University of Hamburg,\\ Hamburg, Germany\\
%\institute{Knowledge Technology, Dept. of Informatics, University of Hamburg\\
\email{\{jan-gerrit.habekost,stefan.wermter\}@uni-hamburg.de}}
\maketitle              % typeset the header of the contribution

\begin{abstract}
The paper introduces CycleIK, a neuro-robotic approach that wraps two novel neuro-inspired methods for the inverse kinematics (IK) task---a Generative Adversarial Network (GAN), and a Multi-Layer Perceptron architecture. These methods can be used in a standalone fashion, but we also show how embedding these into a hybrid neuro-genetic IK pipeline allows for further optimization via sequential least-squares programming (SLSQP) or a genetic algorithm (GA). The models are trained and tested on dense datasets that were collected from random robot configurations of the new Neuro-Inspired COLlaborator (NICOL), a semi-humanoid robot with two redundant 8-DoF manipulators. We utilize the weighted multi-objective function from the state-of-the-art BioIK method to support the training process and our hybrid neuro-genetic architecture. We show that the neural models can compete with state-of-the-art IK approaches, which allows for deployment directly to robotic hardware. Additionally, it is shown that the incorporation of the genetic algorithm improves the precision while simultaneously reducing the overall runtime.

\vspace{-0pt}

\end{abstract}

\begin{figure}[!h]
\vspace{-25pt}
\centering

\begin{minipage}{.36\textwidth}
  \raggedright
   \includegraphics[width=0.98\linewidth]{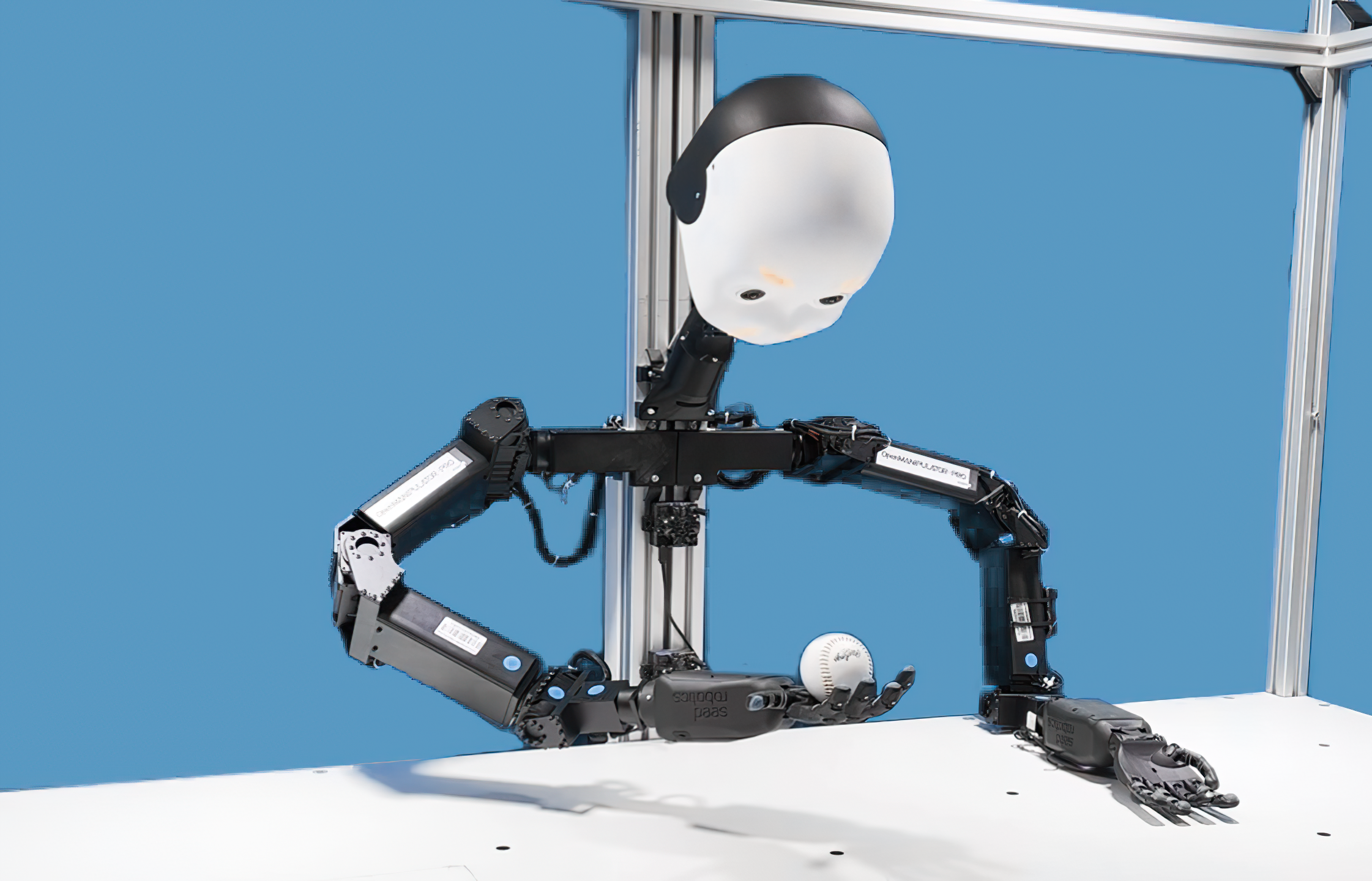}

    \vspace{3pt}

  \includegraphics[width=0.98\linewidth]{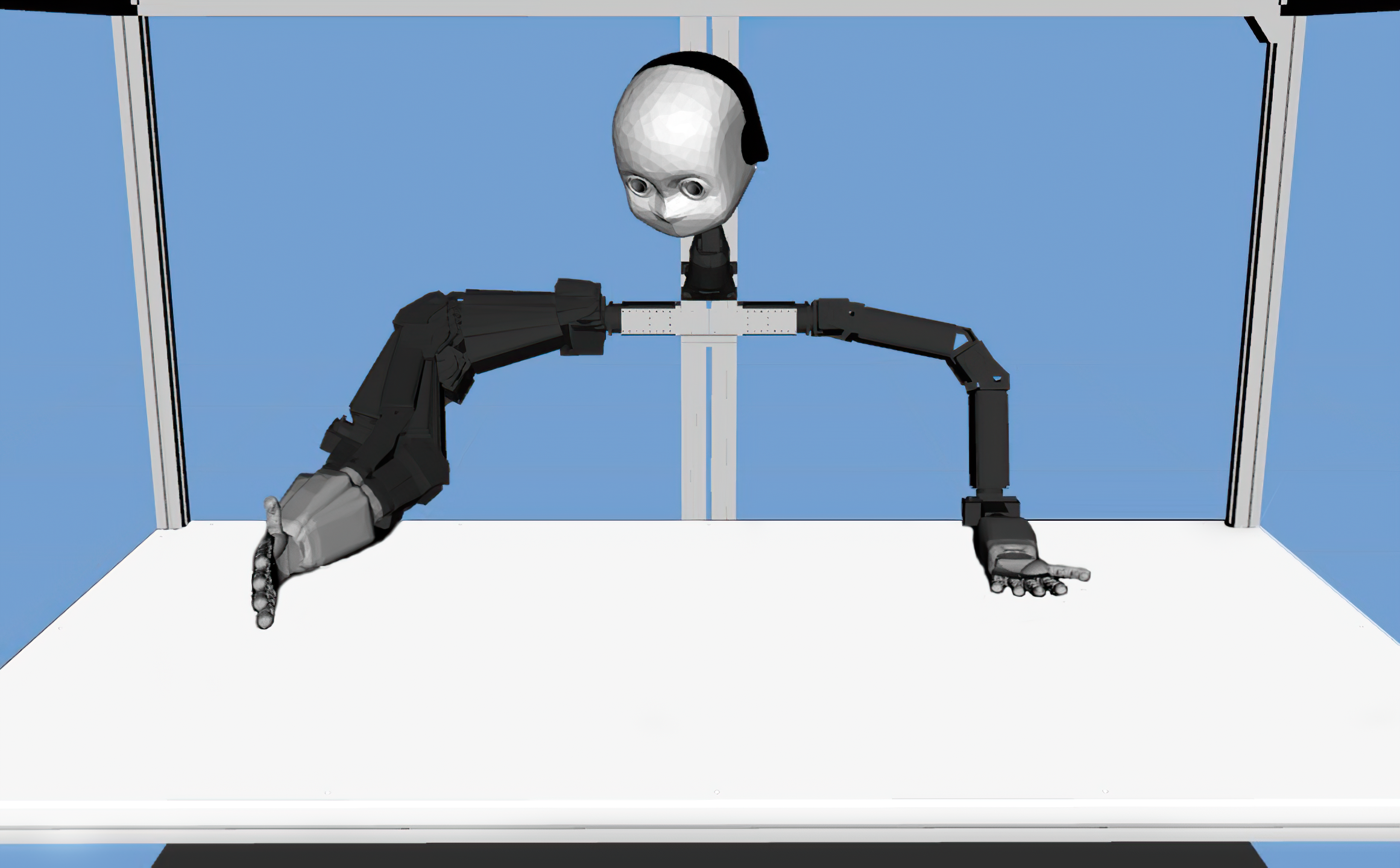}
  
\end{minipage}%
\begin{minipage}{.36\textwidth}
  \raggedleft

   \includegraphics[width=0.98\linewidth]{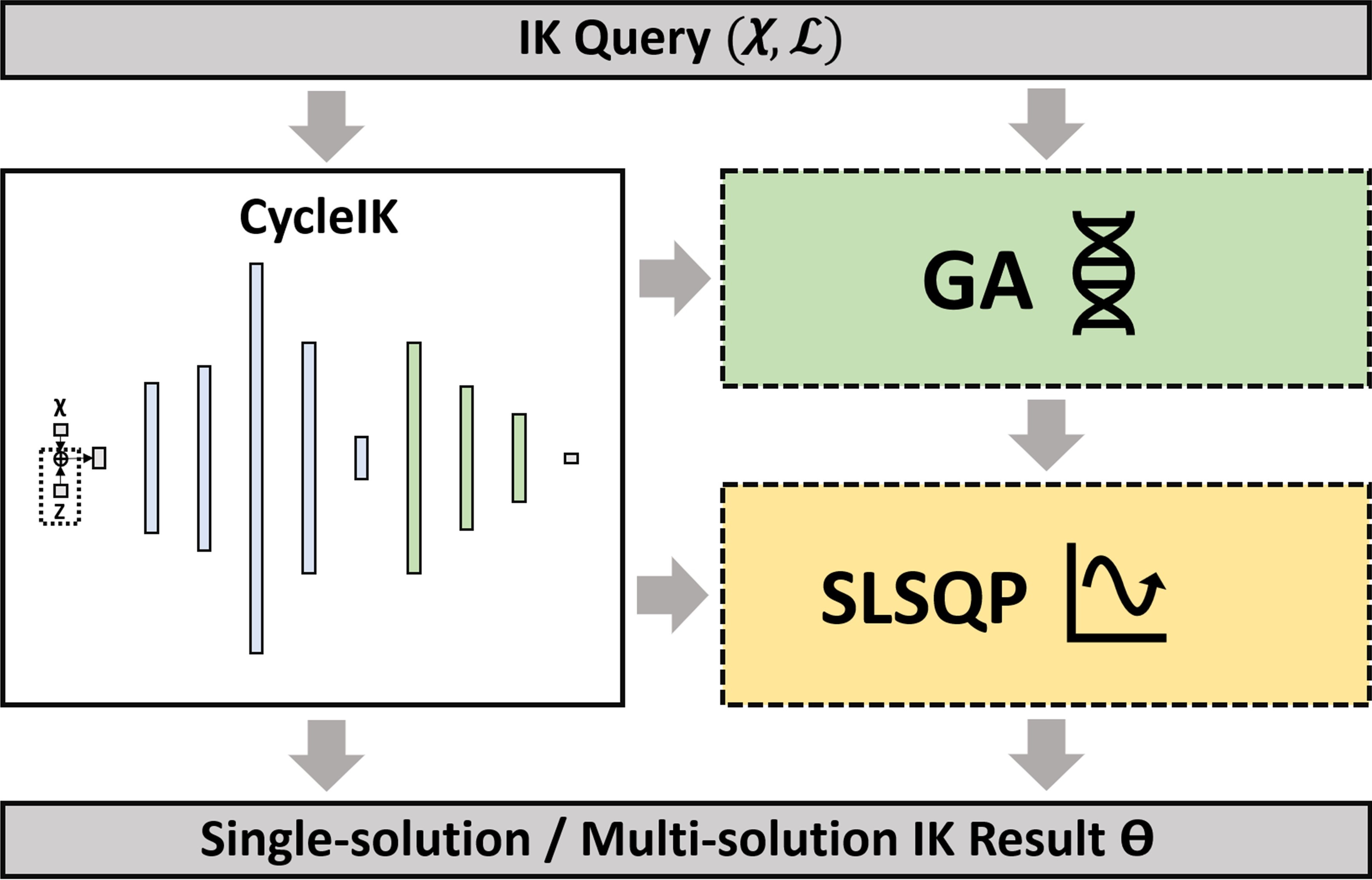}

    \vspace{3pt}
  \includegraphics[width=0.98\linewidth]{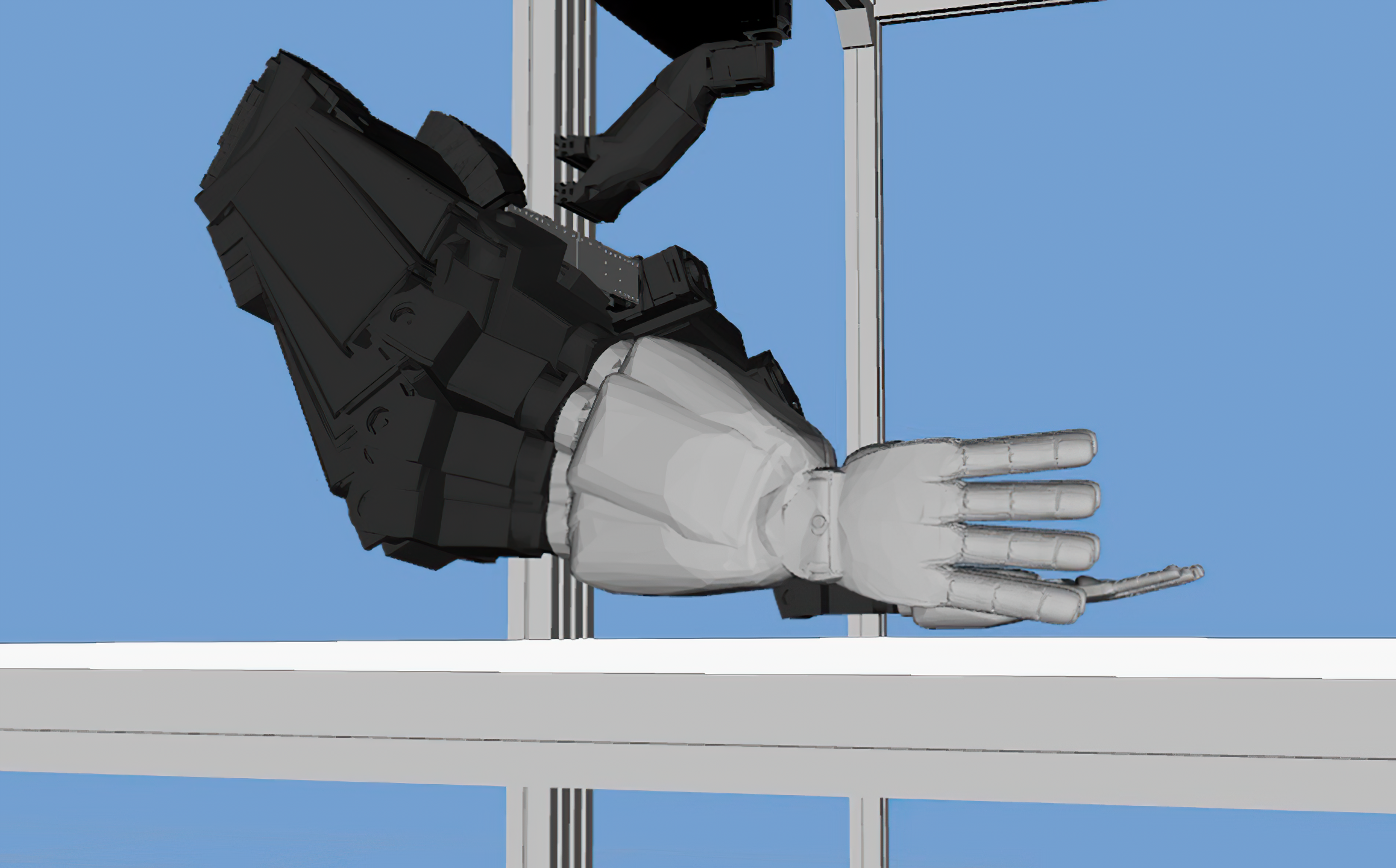}

\end{minipage}

%\begin{minipage}{.86\textwidth}
\vspace{-4pt}
%\begin{minipage}{.825\textwidth}
\caption{CycleIK deployed to physical NICOL hardware (top-left). CycleIK hybrid neuro-genetic inverse kinematics pipeline (top-right). Visualization of the nullspace manifold from the CycleIK Generative Adversarial Network (bottom).}
\label{fig:introduction}
\vspace{-26pt}

\end{figure}

\section{Introduction}
\vspace{-8pt}

The inverse kinematics task searches for suitable joint configurations for a kinematic chain in order to achieve a specified end-effector Cartesian pose. %to fit the Cartesian pose of the end-effector to a specified target.
Recent collaborative and humanoid robot designs often rely on redundant manipulators with more than six degrees of freedom (DoF). The complexity of the inverse kinematics task is therefore increased, as the problem is then under-determined and a set of redundant solutions for a single pose can be found, referred to as the nullspace. The Python-based genetic IK solver Gaikpy \cite{Kerzel2020Neuro-GeneticGrasping}, originally developed for the child-sized NICO robot \cite{Kerzel2017NICO-Neuro-inspiredInteraction} with 6-DoF arms, requires a long runtime in order to deal with the 8-DoF manipulators of the recently developed Neuro-Inspired COLlaborator \cite{kerzel2023nicol}, pictured in the top-left image in \nolinebreak Fig.~\ref{fig:introduction}. 

Traditionally, Jacobian-based methods are utilized for the IK task, such as KDL \cite{SmitsKDL:Library} and Trac-IK \cite{Beeson2015TRAC-IK:Kinematics} which are popular plug-ins in the MoveIt \cite{Coleman2017ReducingStudy} framework and can currently be seen as the industry standard. Both analytical solvers require a high runtime when deployed to NICOL, and have a higher error than Gaikpy \cite{kerzel2023nicol}. We initially configured BioIK \cite{Starke2017AMotion} to be the default solver, a popular state-of-the-art genetic approach, which was also deployed via Moveit. MoveIt, however, does not return a solution for an IK query, when the error is higher than the internal threshold, leaving the control cycle of the robot with no action.

Neural inverse kinematics is a field that unites a wide range of neuro-robotic applications that control the configuration space of a robotic system. The inverse kinematics task is fundamentally embodied in every action-generating neural architecture that takes data from Cartesian space as input. Explicit neural approaches to the task, however, rarely show results with high precision and are distributed over the different application domains of inverse kinematics ranging from robotics to character animation.

Two neural architectures, an auto-regressive Multi-Layer Perceptron (MLP) and a normalizing flow-based Generative Adversarial Network, are proposed in this work. The models solve the inverse kinematics task for a given pose in the reachability space of NICOL and can be deployed directly to robotic hardware, or alternatively be optimized with Gaikpy. The MLP returns exactly one solution for the IK task, while the GAN allows for the exploration of the nullspace manifold. The method is inspired by CycleGAN \cite{Zhu2017UnpairedNetworks}, which trains a dual-GAN architecture in an unsupervised fashion, to transform between two image domains. The positional and rotational errors are measured in Cartesian space by calculating the forward kinematics (FK) for a set of IK solutions that are inferred from the neural models. The FK function calculates the end-effector pose from a given robot configuration and has a short runtime of below $1ms$. Consequently, a second generator as in the original dual-GAN setup of CycleGAN, that approximates the FK function to transform from configuration to Cartesian space, is not necessarily needed for this application. 

\vspace{-13pt}
\section{Related Work}
\vspace{-8pt}

The most similar normalizing flow-based approaches to ours are IKFlow from Ames et al. \cite{Ames2022IKFlow:Solutions} and the work of Kim and Perez \cite{Kim2021LearningManipulator}. IKFlow is a recent and promising neural IK approach. The authors propose a conditional normalizing flow network for the inverse kinematics task, a form of Invertible Neural Network (INN) \cite{Ardizzone2018AnalyzingNetworks}, introduced by Ardizzone et al. for invertible problems. Samples from a simple normal distribution are transformed into valid solution manifolds in the configuration domain through coupling layers that consist of multiple simple invertible functions. The solution manifold can optionally be further optimized with Trac-IK\nolinebreak \cite{Beeson2015TRAC-IK:Kinematics}.  

The approach of Kim and Perez \cite{Kim2021LearningManipulator} has a very similar architecture to IKFlow. Compared to IKFlow, which calculates the error with analytical forward kinematics, Kim and Perez use a second neural network to approximate the FK function in an autoencoder architecture. The approach of Kim and Perez has a comparably high error in the centimeter range and requires further optimization with the Jacobian, while IKFlow reaches a millimeter range of error.

Lembono et al. \cite{Lembono2020LearningNetwork} present an ensemble architecture in which multiple GAN generators learn to sample from disjunct patches of the configuration space. A single forward kinematics discriminator is used that also checks for further constraints, e.g. minimal displacement of the arms. A more detailed investigation of GANs in the context of IK is given by Ren and Ben-Tzvi \cite{Ren2020LearningNetworks}. The paper modifies four different types of GAN architectures to solve the inverse kinematics problem. The discriminator produces binary output, while most GAN designs perform regression and calculate the continuous error to the target pose.

Bensadoun et al. \cite{Bensadoun2022NeuralKinematic} introduce a Gaussian Mixture Model (GMM) ensemble to calculate multiple solutions for the IK problem. A GMM is created for every joint in the kinematic chain. A hypernet parameterizes the GMMs conditioned to the target pose. Volinski et al. \cite{Volinski2022Data-drivenNeurorobotics} utilize Spiking Neural Networks (SNN) to solve the inverse kinematics problem. The approach trains three different variations of simple SNN architectures. ProtoRes \cite{Oreshkin2021ProtoRes:Kinematics} was introduced by Oreshkin et al. to reconstruct natural body poses from sparse user input for animation tasks. The framework consists of a pose encoder that creates a latent embedding from the user input and then solves the IK task with a pose decoder. 

\vspace{-10pt}
\section{Method}
\vspace{-8pt}

\begin{wrapfigure}{r}{0.5\textwidth}
\vspace{-32pt}
  \begin{center}
    \includegraphics[width=0.5\textwidth]{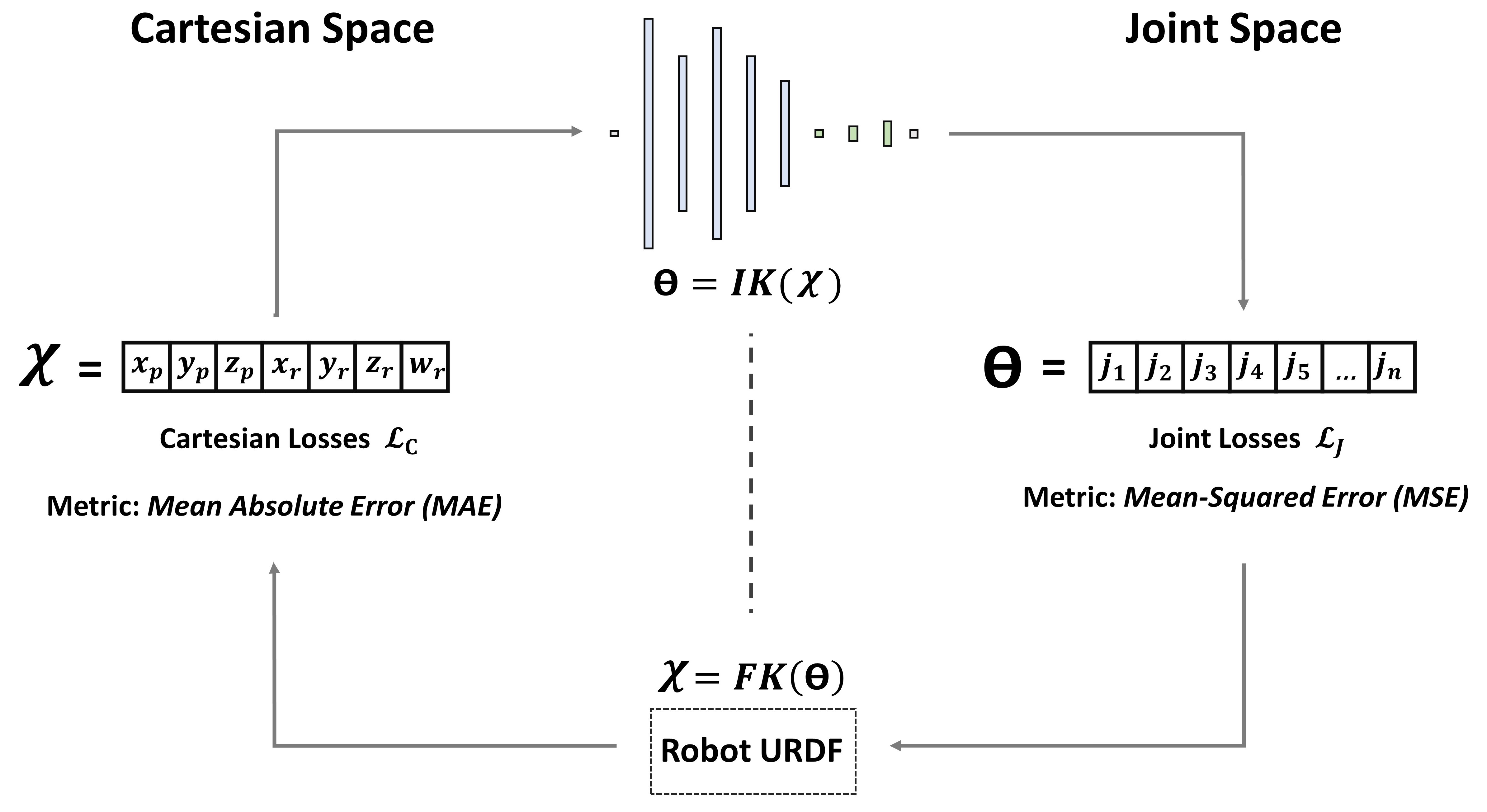}

  \end{center}
  \vspace{-15pt}
  \caption{CycleIK neuro-inspired training and architecture overview. A batch of Cartesian poses $\mathcal{X}$ is inferred by the network to predict a set of valid robot configurations $\Theta$ under constraints $\mathcal{L}$.}
   \label{fig:cycleik_oveview}
  \vspace{-25pt}
\end{wrapfigure}

We propose CycleIK, a neuro-inspired inverse kinematics solver that makes use of the cyclic dependency between the transformation from configuration to Cartesian space and its inverse. An overview of the architecture is given in Fig.~\ref{fig:cycleik_oveview}. The framework enables either training a single-solution auto-regressive Multi-Layer Perceptron or a normalizing flow-based GAN architecture that allows the parallel inference of multiple redundant solutions within 1$ms$. Furthermore, the approach can be utilized as a neuro-kinematic toolbox. The default networks can be substituted by any end-to-end or multi-stage robotic control architecture that predicts joint angles and provides a Cartesian pose as a label. CycleIK is implemented in PyTorch, to be as openly available as possible.
%, while most popular IK solvers are implemented in C++. 
Most IK solvers are implemented in C++ and generally rely on iterative numerical methods for the optimization process, often leading to a higher runtime compared to the inference of a neural network.

CycleIK treats the joint space as a semi-hidden domain, and calculates positional and rotational losses only in Cartesian space, by inferring a full cycle back to Cartesian space, as shown in the following equations (Eq. 1 and 2):
\vspace{-7pt}
\begin{align}
\hat{\mathcal{X}} &= FK(IK(\mathcal{X})) \\
e_{IK} &= \|\hat{\mathcal{X}} - \mathcal{X}\|
\end{align}

\vspace{-5pt}
%$x_i$
Where $\mathcal{X}$ is a batch of an arbitrary natural number of target poses, and $e_{IK}$ is the linear Cartesian error. While learning a one-to-one mapping between data from Cartesian space and corresponding joint angles $\theta$ can work for lower-DoF manipulators \cite{Kerzel2020Neuro-GeneticGrasping}, the approach shows a high error for redundant manipulators like on the NICOL robot \cite{kerzel2023nicol}, as these manipulators have a one-to-many mapping in the form of the redundant nullspace manifold $\Theta$. Thus, we minimize the linear Cartesian error $e_{IK}$ instead, which in our experience learns and generalizes more smoothly.

\begin{figure}[!b]
\vspace{-10pt}
\centering
\begin{minipage}{.5\textwidth}
  \centering
   \includegraphics[width=0.94\linewidth]{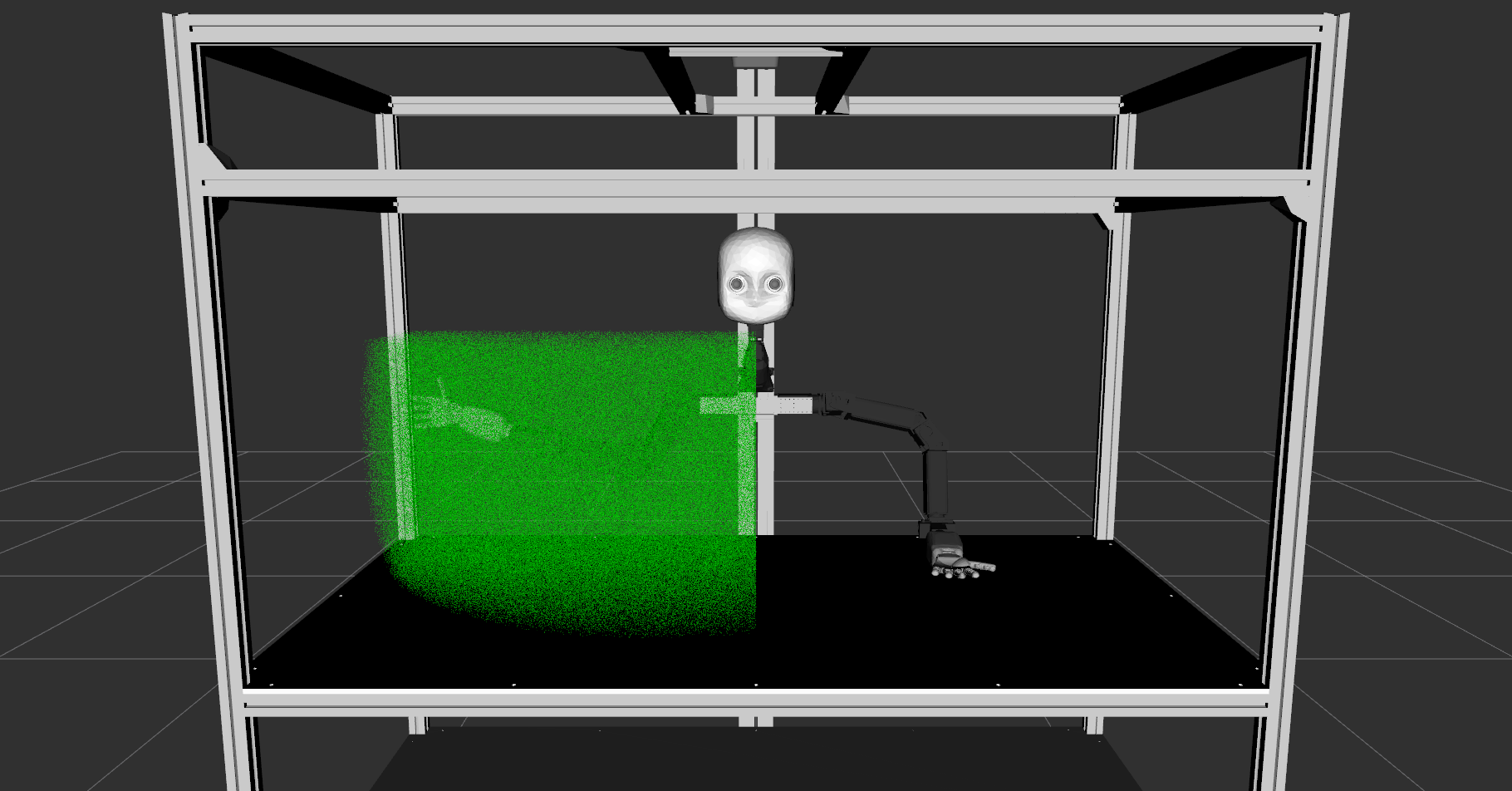}
  $Small_{1000}$ dataset front view
  %\vspace{5pt}
  
\end{minipage}%
\begin{minipage}{.5\textwidth}
  \centering

   \includegraphics[width=0.94\linewidth]{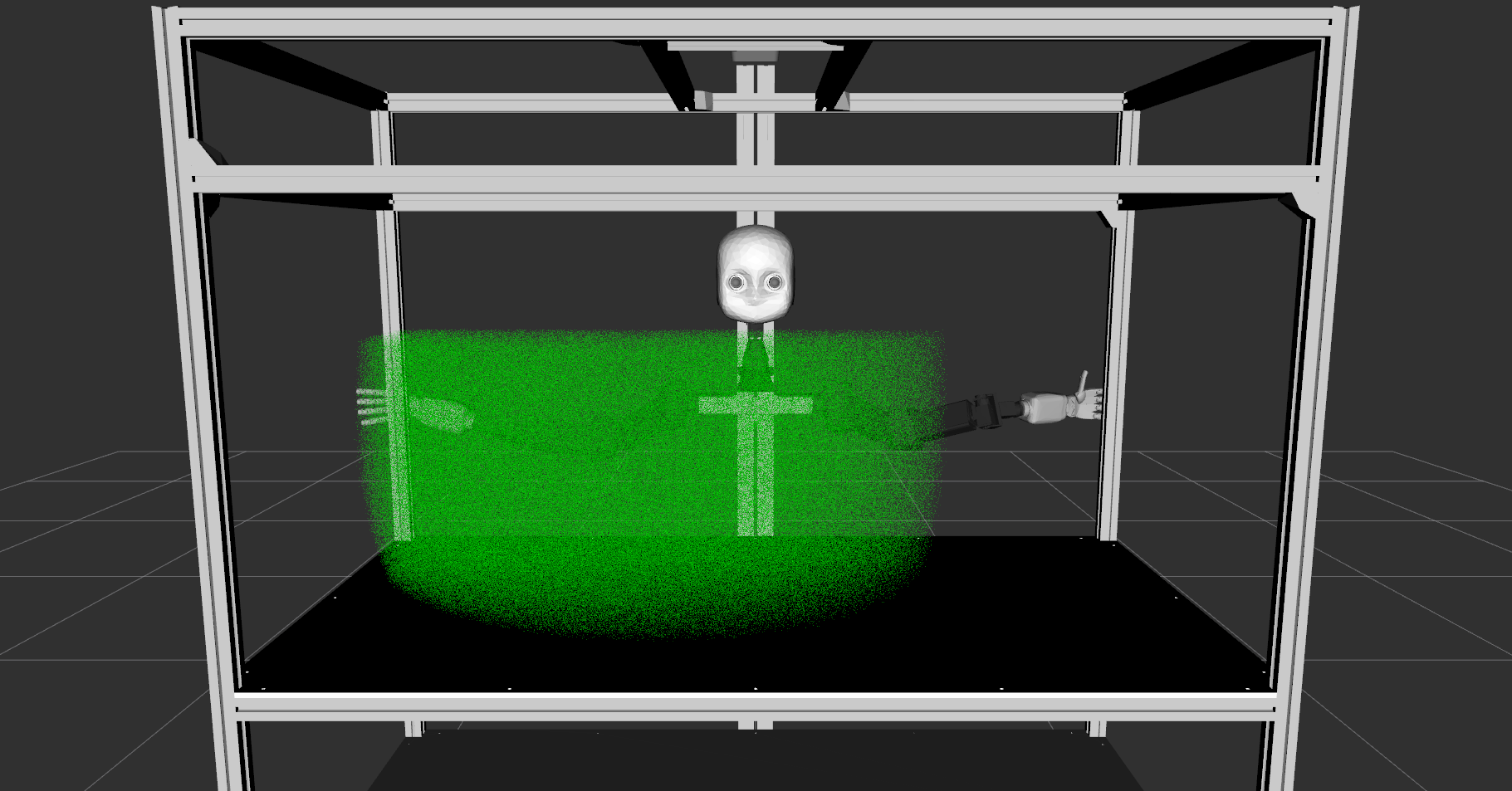}
  $Full_{1400}$ dataset front view

\end{minipage}
\vspace{-5pt}
\caption{Visualization of NICOL's right arm workspace, with the $Small_{1000}$ dataset on the left and the $Full_{1400}$ dataset on the right.}
%\vspace{-20pt}
\label{fig:workspace_viz}
\end{figure}

Similar to neuro-inspired multi-solution solvers like IKFlow and CycleIK, genetic algorithms produce multiple solutions for an IK query, and have shown good results for the IK task \cite{Kerzel2020Neuro-GeneticGrasping,Starke2017AMotion,Aguilar2011InverseAlgorithm}. The most popular genetic IK approach is BioIK \cite{Starke2017AMotion}, which is available in both MoveIt and Unity. The method supports genetic algorithms by hybridization with particle swarm optimization (PSO). The architecture allows generic IK queries through a weighted partial cost function $\phi(\Theta, \mathcal{L})$ that is applied to the set of IK solutions $\Theta$ under the constraints $\mathcal{L}$. The constraints can be reformulated at every IK query, so complex dynamic tasks such as collision avoidance in motion planning can be performed. Different goal types can be set for either the links or joints of the robot. We adapt the weighted partial cost function from BioIK for both of our models. CycleIK's single-solution model optionally makes use of a set of weighted constraints $\mathcal{L} = \mathcal{L}_C \cup \mathcal{L}_J$, that consists of specified goals, either in Cartesian or joint space. The constraints are applied by the multi-objective function in every training step. Both, the single-solution MLP as well as the multi-solution GAN, can optionally be further optimized by the Python-based genetic IK Gaikpy \cite{Kerzel2020Neuro-GeneticGrasping} or non-linear sequential least squares quadratic programming \cite{Kraft1988AProgramming}, where again a partial weighted cost function can be used to select the optimal solution. An overview of the neuro-genetic IK pipeline is given in the top-right image of Fig.~\ref{fig:introduction}.

\vspace{-10pt}
\subsection{Dataset}
\vspace{-5pt}
Three datasets were collected from NICOL's workspace: $Small_{1000}$, $Full_{1000}$ and $Full_{1400}$. Uniform random collision-free robot configurations were sampled. The $Small_{1000}$ and $Full_{1400}$ dataset are shown in Fig.~\ref{fig:workspace_viz}. The $Small_{1000}$ dataset contains 1,000,000 samples and is limited to the right side of the tabletop, which is located 80cm above the ground. The $Full_{1000}$ and $Full_{1400}$ datasets with $1,000,000$ and $1,400,000$ poses are sampled from the whole workspace of the right arm over the tabletop. We built test sets with 10\% size and validation sets with 1\% size for each of the training datasets. In all datasets, a 20cm safety margin was included at the back of the workspace on the x-axis, as well as a 10cm safety margin on the y-axis on the right-hand side of the robot workspace. All properties of the datasets can be seen in Table~\ref{tb:dataset}. A convex hull was generated around the data points to approximate the Cartesian volume of each dataset.

\begin{table}
\vspace{-20pt}
\centering
\caption{Overview of the training datasets and the corresponding $10\%$ test sets ($Small_{100}$, $Full_{100}$ and $Full_{140}$) and $1\%$ validation sets ($Small_{10}$, $Full_{10}$ and $Full_{14}$).}
\vspace{10pt}
\resizebox{0.6\columnwidth}{!}{
\begin{tabular}[!]{||c|cccc||}
\hline
\multirow{2}{*}{\textbf{Dataset}}&\textbf{Workspace}& \multirow{2}{*}{\textbf{Samples}}&\textbf{Volume}&\textbf{Sample Density}  \\ [0.5ex]
&[x, y, z] ($m$)& &\textbf{($cm^3$)}&\textbf{($samples$ $per$ $cm^3$)}  \\ [0.5ex]
\hline\hline
  \multirow{2}{*}{$Small_{1000}$} & & \multirow{2}{*}{$10^6$} & \multirow{2}{*}{$295.56 \cdot 10^3$} & \multirow{2}{*}{$3.383$}\\
   &  & & &\\
   %\hline
  \multirow{2}{*}{$Small_{100}$} & \multirow{2}{*}{$\left[\begin{array}{ccc}
0.2 & -0.9 & 0.8 \\
0.85 & 0.0 & 1.4 \end{array}\right]$} & \multirow{2}{*}{$10^5$} & \multirow{2}{*}{$293.34 \cdot 10^3$} & \multirow{2}{*}{$0.341$}\\
   & & & &\\
   %\hline
  \multirow{2}{*}{$Small_{10}$} &  & \multirow{2}{*}{$10^4$} & \multirow{2}{*}{$287.11 \cdot 10^3$} & \multirow{2}{*}{$0.035$}\\
   & & & &\\
   \hline\hline
  \multirow{2}{*}{$Full_{1000}$} &  & \multirow{2}{*}{$10^6$} & \multirow{2}{*}{$420.11 \cdot 10^3$} & \multirow{2}{*}{$2.38$}\\
   & & & &\\
   %\hline
  \multirow{2}{*}{$Full_{100}$} &  \multirow{2}{*}{$\left[\begin{array}{ccc}
0.2 & -0.9 & 0.8 \\
0.85 & 0.48 & 1.4 \end{array}\right]$} & \multirow{2}{*}{$10^5$} & \multirow{2}{*}{$415.13 \cdot 10^3$} & \multirow{2}{*}{$0.241$}\\
   &  & & &\\
   %\hline
  \multirow{2}{*}{$Full_{10}$} &  & \multirow{2}{*}{$10^4$} & \multirow{2}{*}{$401.75 \cdot 10^3$} & \multirow{2}{*}{$0.025$}\\
   &  & & &\\
   \hline\hline
  \multirow{2}{*}{$Full_{1400}$} & & \multirow{2}{*}{$1.4 \cdot 10^6$} & \multirow{2}{*}{$420.43 \cdot 10^3$} & \multirow{2}{*}{$3.33$}\\
   & & & &\\
   %\hline
  \multirow{2}{*}{$Full_{140}$} &\multirow{2}{*}{$\left[\begin{array}{ccc}
0.2 & -0.9 & 0.8 \\
0.85 & 0.48 & 1.4 \end{array}\right]$} & \multirow{2}{*}{$1.4 \cdot 10^5$} & \multirow{2}{*}{$416.09 \cdot 10^3$} & \multirow{2}{*}{$0.336$}\\
   & & & &\\
   %\hline
  \multirow{2}{*}{$Full_{14}$} & & \multirow{2}{*}{$1.4 \cdot 10^4$} & \multirow{2}{*}{$405.07 \cdot 10^3$} & \multirow{2}{*}{$0.035$}\\
   &  & & &\\
   \hline
\end{tabular}
}
\label{tb:dataset}
\vspace{-10pt}
\end{table}

\vspace{-15pt}

\begin{figure}[h]
\vspace{-5pt}
\centering
    \includegraphics[width=0.99\linewidth]{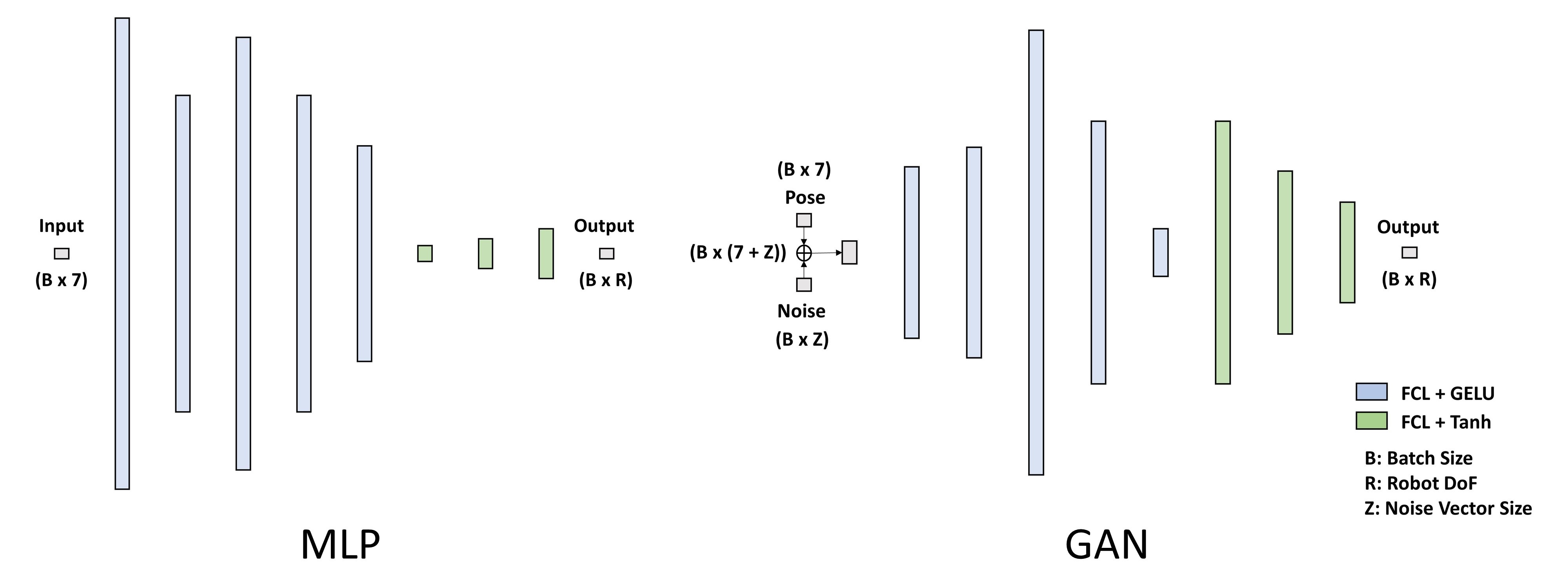}
    \vspace{-10pt}
    \caption{Neural architectures optimized for the $Small_{1000}$ dataset, Multi-Layer Perceptron (left) and Generative Adversarial Network (right).}
    \label{fig:network_viz}
\vspace{-15pt}
\end{figure}

%(see Fig. \ref{fig:cycleik_oveview})
\subsection{Architecture}
\vspace{-4pt}
The basic network architecture is very similar for both models. The pose is encoded as a 7-dimensional vector, i.e. the 3-dimensional position $[x_p, y_p, z_p]^T$ concatenated with the rotation represented as a 4-dimensional unit quaternion $[x_r, y_r, z_r, w_r]^T$, as shown in Fig. \ref{fig:cycleik_oveview}. The output of the network has the same dimension as the robot DoF, so every field of the output vector corresponds to a motor position in the kinematic chain. The GAN additionally concatenates the pose with a second input, a random uniform noise vector that is utilized to sample from the nullspace manifold. The models utilize two different activation functions. While Gaussian-Error Linear Units \cite{Hendrycks2016GaussianGELUs} (GELU) are generally used for all the layers, the Tanh activation is applied to the last one to three layers of the network, as this highly improves the results. The data is normalized to lie in the interval $[-1, 1]$, which is equivalent to the limits of the network input and output. Thus, the method cannot push the joint angles through their joint limits, which is a shortcoming of a lot of Jacobian-based IK solvers. Visualizations of the two network architectures for the NICOL robot can be found in Fig.~\ref{fig:network_viz}.

\vspace{-10pt}

\subsection{Training}
\vspace{-5pt}
In every training step, a batch of poses $\mathcal{X}$ is inferred by the network. For the single-solution network, the training step is straightforward---after inference, forward kinematics are applied to the batch of solutions $\Theta$ to determine the reached poses $\hat{\mathcal{X}}$ and then apply the multi-objective loss function, as in Eq. \ref{eq:multi_objective_loss}:
\vspace{-5pt}
\begin{equation}
    loss_{\mathcal{L}} = \phi(\Theta, \hat{\mathcal{X}}, \mathcal{L})
    \label{eq:multi_objective_loss}
    \vspace{-5pt}
\end{equation}

Here, $\mathcal{L}$ holds at least the positional and rotational error. For the NICOL robot, we applied a zero-controller goal that minimizes the displacement of the motor position from the zero position of the selected subset of redundant joints in the kinematic chain. Our preliminary experiments showed the best performance by using the mean absolute error for Cartesian space losses and mean squared error for the joint space losses, as the error increased for all other evaluated error terms. The learning rate is decreased linearly at the end of each epoch.

The training process for the multi-solution GAN extends the training process of the MLP. After calculation of the positional and rotational loss for a batch of Cartesian samples from the training set, one of the poses is randomly chosen from the batch. A tensor of the same size as the training batch is created and filled with the chosen pose. Random uniform noise $\mathcal{Z}$ of the required batch size and noise vector size is then generated and used for the forward pass. The training aims to maximize the variance in the solution batch $\Theta$. The normalizing flow method is applied, as the network is not being forced to regress to only one solution, but instead fit the nullspace distribution $\Theta$ to the noise $\mathcal{Z}$, as in Eq. \ref{eq:variance_loss_term}: 

\vspace{-9pt}
\begin{equation}
    loss_{var} = MSE(var(\Theta) - var(\mathcal{Z}))
    \vspace{-2pt}
    \label{eq:variance_loss_term}
\end{equation}

The method can produce multiple valid solutions for the NICOL robot with millimeter-level accuracy. One possible extension would be to combine Kullback-Leibler divergence for the loss and normally distributed noise in the input, as done by IKFlow \cite{Ames2022IKFlow:Solutions} and Kim and Perez \cite{Kim2021LearningManipulator}.

\vspace{-10pt}

\subsection{Optimization}
\vspace{-5pt}
Each of the models was optimized over 250 trials for both the Small and Full workspace. The results are shown in Table~\ref{tb:optuna_result}. For the Full workspace, we chose to optimize the models with the $Full_{1400}$ dataset. We used the Optuna framework \cite{Akiba2019Optuna:Framework} to optimize the models with a Tree-structured Parzen Estimator (TPE) for sampling, and a hyperband pruner. Four parameters were defined for the optimization process, which are the batch size, learning rate, number of layers in the network, and the number of layers with tanh activations at the end of the network. Additionally, we optimized the number of neurons in every layer. An overview of the exact network layouts can be found in Table~\ref{tb:optuna_result_nets}, and a visualization of the network structures for the Small workspace is shown in Fig.~\ref{fig:network_viz}. For the GAN only, we also optimized the size of the input noise vector.   

\vspace{-10pt}

\begin{table}[h!]
\centering
\vspace{-10pt}
\caption{Training parameters for the different network types, optimized for the $Small_{1000}$ and $Full_{1400}$ datasets.}
\vspace{10pt}
\resizebox{0.7\columnwidth}{!}{
\begin{tabular}{||c|cc|cc|cc||}
\hline
\multirow{2}{*}{\textbf{Parameter}}&\multicolumn{2}{c}{\textbf{MLP}}& \multicolumn{2}{c}{\textbf{GAN}} &  
\textbf{Parameter} & \textbf{Limits} \\ [0.5ex]
   & Small & Full & Small & Full & min. / max. & step size\\ [0.5ex]
 \hline\hline
 \multirow{2}{*}{Batch Size}  & \multirow{2}{*}{$150$} & \multirow{2}{*}{$300$} & \multirow{2}{*}{$350$} & \multirow{2}{*}{$300$} & \multirow{2}{*}{$100$ / $600$}& \multirow{2}{*}{$50$} \\ 
  &   &  &  &  &   &\\ 
  \hline
 
\multirow{2}{*}{Learning Rate}  & \multirow{2}{*}{$1.6\cdot 10^{-4}$} & \multirow{2}{*}{$10^{-4}$} & \multirow{2}{*}{$2.1\cdot 10^{-4}$} & \multirow{2}{*}{$1.9\cdot 10^{-4}$} & \multirow{2}{*}{$10^{-5}$ / $10^{-3}$}& \multirow{2}{*}{$10^{-5}$} \\ 
  &   &  &  &  &   &\\ 
  \hline
  \multirow{2}{*}{Number Layers}  & \multirow{2}{*}{$8$} & \multirow{2}{*}{$8$} & \multirow{2}{*}{$8$} & \multirow{2}{*}{$8$} & \multirow{2}{*}{$7$ / $9$}& \multirow{2}{*}{$1$} \\ 
  &   &  &  &  &   &\\ 
  \hline
  Number   & \multirow{2}{*}{$3$} & \multirow{2}{*}{$3$} & \multirow{2}{*}{$3$} & \multirow{2}{*}{$2$} & \multirow{2}{*}{$1$ / $3$}& \multirow{2}{*}{$1$} \\ 
  Tanh Layers&   &  &  &  &   &\\ 
  \hline
  Noise Vector  & \multirow{2}{*}{$\textbf{-}$} & \multirow{2}{*}{$\textbf{-}$} & \multirow{2}{*}{$8$} & \multirow{2}{*}{$10$} & \multirow{2}{*}{$3$ / $10$}& \multirow{2}{*}{$1$} \\ 
  Size&   &  &  &  &   &\\ 
  \hline
\end{tabular}
}
\label{tb:optuna_result}
\vspace{-25pt}
\end{table}

\vspace{5pt}
\begin{table}[h!]
%\vspace{-20pt}
\centering
\vspace{-20pt}
\caption{Network structures of the different network types optimized for the $Small_{1000}$ and $Full_{1400}$ workspace.}
\vspace{10pt}
\resizebox{0.7\columnwidth}{!}{
\begin{tabular}[H!]{||cc|c||}
\hline
\textbf{Model} & \textbf{Workspace}&\textbf{Neurons per Layer}\\
\hline
\multirow{2}{*}{MLP} & Small & $[3380, 2250, 3240, 2270, 1840, 30, 60, 220]$\\
& Full & $[2200, 2400, 2400, 1900, 250, 220, 30, 380]$\\
 \hline
\multirow{2}{*}{GAN} & Small & $[ 790, 990, 3120, 1630, 300, 1660, 730, 540 ]$\\
& Full & $[ 1180, 1170, 2500, 1290, 700, 970, 440, 770 ]$\\
 \hline
\end{tabular}}
\label{tb:optuna_result_nets}
\vspace{0pt}
\end{table}

\vspace{-27pt}
\section{Results}
\vspace{-7pt}
The application of the weighted partial cost function on the MLP network and the variance loss on the GAN created stability issues in the training process of differing severity for the two models. The MLP rarely shows stability issues during the training process, but they sometimes occur when trained for more than 100 epochs, and can be dealt with using gradient clipping. The GAN suffers more severe stability issues, and could not be trained for more than 50 epochs in our experiments. We hypothesize it is due to the competition of maximizing the nullspace manifold variance while maintaining precise IK regression. Gradient clipping cannot be applied as easily in the case of the GAN because it prevents learning proper minimization of the variance loss.

\begin{figure}[t!]
    \vspace{-5pt}
  \centering
    \includegraphics[width=0.98\textwidth]{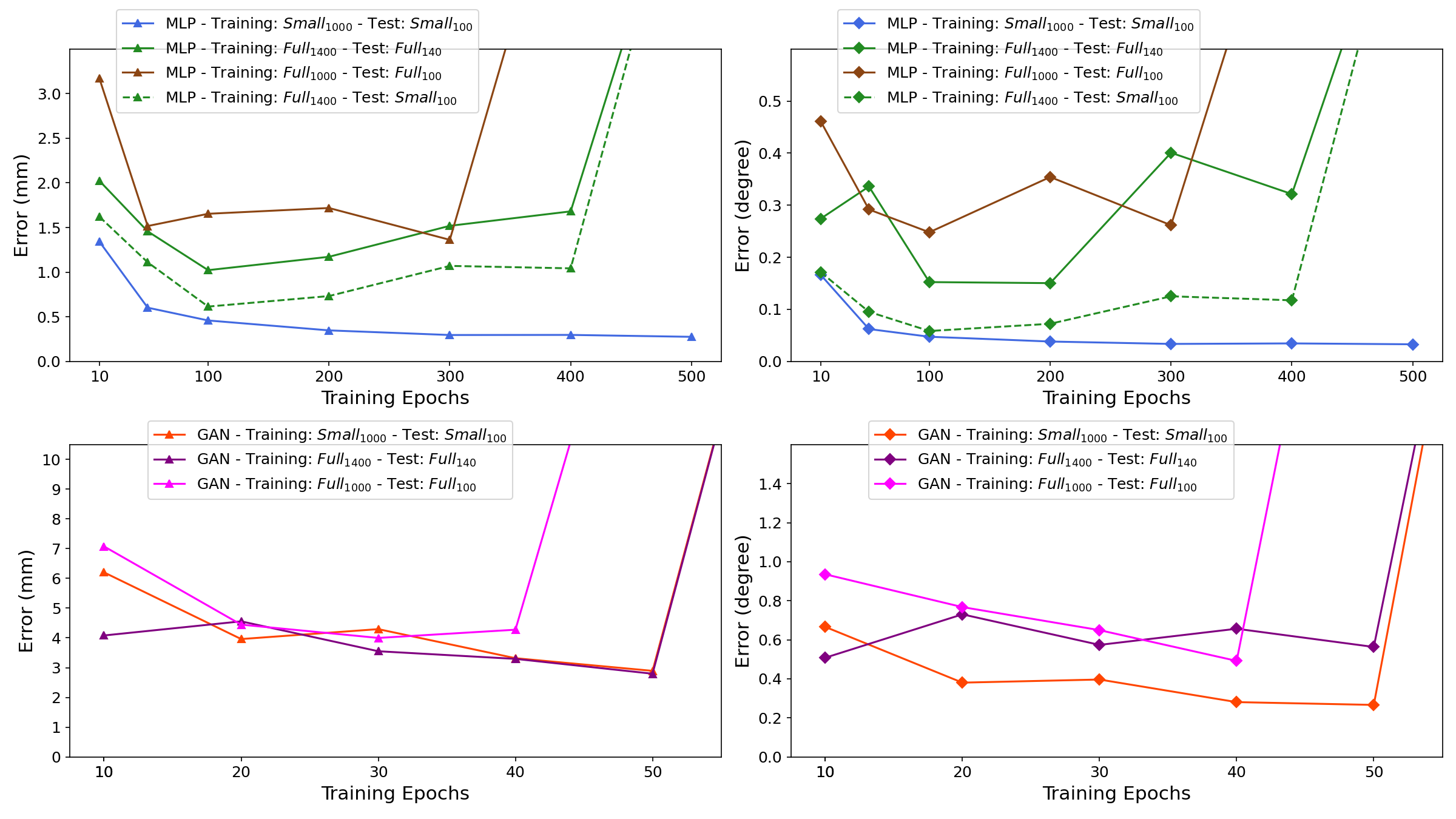}
    \vspace{-8pt}
\caption{Average positional and rotational error of the MLP and GAN model under training for varying numbers of epochs.}
\label{fig:results_over_epochs}
\vspace{-15pt}
\end{figure}

\vspace{-10pt}
\subsection{Optimal Number of Epochs}
\vspace{-5pt}
To determine the optimal number of epochs for the training process, we trained both presented models for each of the three datasets that were generated, so that six models in total were evaluated under different epoch configurations. A standalone training was performed for every individual model and number of epochs. To handle the stability issues of the models, we gave every evaluated epoch configuration a number of restarts in case stability issues occur. Each choice of maximum epochs was allowed two restarts for the MLP and nine for the GAN. If exploding gradients occurred in every observed training, the combination was considered to have failed. The results of our experiments are shown in Fig.~\ref{fig:results_over_epochs}. We calculated the positional and rotational error for the MLP by first taking the average over the three corresponding axes of the 6-DoF pose error, and then averaging the results for the whole 10\% test sets. For the multi-solution GAN, we first calculated the average error over single batches of nullspace solutions, before taking the mean over the whole test set. We take the success definition for the inverse kinematics task from Kerzel et al. \cite{Kerzel2020Neuro-GeneticGrasping}, which allows $10mm$ positional and $20$-degree rotational error.

\textbf{GAN} It can be seen that the training process of the $Small_{1000}$ dataset had the lowest error for most of the epoch configurations when compared to the training of the $Full_{1000}$ and $Full_{1400}$ datasets. For a higher number of epochs, the positional error of the GAN models behaves similarly for the $Small_{1000}$ and $Full_{1400}$ datasets. Instabilities occur for short training and with regard to the rotational error. The training of the $Full_{1000}$ GAN model already starts to fail when training for more than 40 epochs, while the rotational error can compete with the loss of the $Full_{1400}$ model for a lot of configurations.

\textbf{MLP} The results of the single-solution MLP for the training on the $Small$ and $Full$ workspace differ more strongly than for the GAN. Different from the GAN, where the exact same loss is used for both workspaces, the zero-controller goal that we set for the training of the MLP has to be tuned for a specific workspace and therefore differs. The additional joint space goal can therefore explain the differences in training behavior to some degree. The training with the $Full_{1000}$ dataset can also for the MLP compete with the $Full_{1400}$ training for some epoch configurations. Overall, the best model for the $Full_{1400}$ dataset exceeds the best model for the $Full_{1000}$ dataset. 

The training with the $Small_{1000}$ dataset proceeded the smoothest, and we did not experience any stability issues. In contrast to the GANs, where the smallest positional error is achieved after 50 epochs for both workspaces, with slightly below 3$mm$ average error, the MLP models differ in the ideal training length as well as in the smallest error. While the lowest positional error for the Full workspace is achieved after 100 epochs, the best results for the Small workspace are found after 300 training epochs. The best results for the $Full_{1000}$ dataset are also achieved with 300 training epochs, but cannot compete with the best model of the $Full_{1400}$ dataset. We evaluated the performance of the different models for the $Full_{1400}$ dataset on the $Small_{100}$ test set to make the $Small_{1000}$ and $Full_{1400}$ models directly comparable. It can be seen from the green dotted line in Fig.~\ref{fig:results_over_epochs} that the $Full_{1400}$ models perform very similarly when evaluated on the same test data as the $Small_{1000}$ models. Positional and rotational errors only show small differences until 100 training epochs are exceeded.

Overall, we focus more on the positional error rather than the rotational error, as the rotational error is far below our success limit of 20 degrees in almost all cases. Especially the models that were only trained for 10 or 20 epochs can show up to $5 mm$ average positional error, and therefore a lot of solutions around the upper bound of the error exceed the limit of $1 cm$.

\vspace{-10pt}
\subsection{Precision Analysis}
\vspace{-5pt}
From the previous experiment, the best-performing models were selected and evaluated for the $Small_{100}$ and $Full_{140}$ test sets. We seeded 50\% of Gaikpy's initial population with solutions from the neural models in a follow-up experiment and filled the other half of the population with uniform random robot configurations within the joint limits. As a baseline, errors for standalone BioIK and Gaikpy were evaluated. The results of our IK experiments on the Small and Full workspace can be seen in Table~\ref{tb:final_results}. The framework offers SLSQP for further optimization but we did not use it in the experiments, as the solutions are already precise enough to be deployed to the physical hardware. 

The performance of BioIK on the test sets was measured via MoveIt. Since MoveIt reports an exception for solutions whose error lies over a specified threshold, as this threshold cannot be influenced, no solution can be evaluated for the failed requests. This behavior is different from all other methods that are utilized in this work, as they always report at least some kind of solution. For the failed MoveIt requests, we calculated the distance between the initial rest end-effector pose and the target pose, which increases the average positional \nopagebreak and rotational error in comparison to the other methods. The positional error lies around $0.02$ to $0.05 mm$ for the successful requests and would therefore outperform the presented Python-based methods. 

\begin{table}[b!]
\vspace{-20pt}
\centering
\caption{Results of different CycleIK variants and standalone Gaikpy and BioIK on the $Small_{100}$ and $Full_{140}$ test set.}
\vspace{10pt}
\resizebox{0.675\columnwidth}{!}{
\begin{tabular}[!b]{||cc|cc|cc|cc||}
\hline
\multirow{2}{*}{\textbf{Model}}&\textbf{Work-}&\multicolumn{2}{c}{\textbf{Position} (mm)}& \multicolumn{2}{c}{\textbf{Orientation} ($^{\circ}$)} &  
\textbf{Success} & \textbf{Timeout} \\ [0.5ex]
  & \textbf{space} & Avg. & Min. / Max. & Avg. & Min. / Max. & \textbf{Rate ($\%$)} &\textbf{($ms$)}\\ [0.5ex]
 \hline\hline
 \multirow{4}{*}{CycleIK$_{MLP}$} & \multirow{2}{*}{Small} & \multirow{2}{*}{$0.295$} & $5.36 \cdot 10^{-4}$ / & \multirow{2}{*}{$0.033$} & $2.39 \cdot 10^{-4}$ / & \multirow{2}{*}{$99.48$}& \multirow{2}{*}{$0.242$} \\ 
  &  &  & $143.56$ &  & $85.73$ &  &\\ 
 
 & \multirow{2}{*}{Full} & \multirow{2}{*}{$1.022$} & $1.12 \cdot 10^{-3}$ / & \multirow{2}{*}{$0.152$} & $3.92 \cdot 10^{-4}$ / &\multirow{2}{*}{$98.49$}& \multirow{2}{*}{$0.243$} \\ 
  &  &  & $376.39$ &  & $127.41$ & & \\ 
 \hline\hline
 \multirow{3}{*}{CycleIK$_{MLP}$} & \multirow{2}{*}{Small} & \multirow{2}{*}{$0.074$} & $4.83 \cdot 10^{-6}$ / & \multirow{2}{*}{$0.089$} & $2.74 \cdot 10^{-4}$/ & \multirow{2}{*}{$99.85$}&\multirow{2}{*}{$19.589$} \\ 
 \multirow{3}{*}{w. Gaikpy} &  &  & $245.57$ &  & $93.43$ & &  \\ 
 
 & \multirow{2}{*}{Full} & \multirow{2}{*}{$0.163$} & $3.44 \cdot 10^{-6}$ / & \multirow{2}{*}{$0.308$} & $1.59 \cdot 10^{-4}$ / & \multirow{2}{*}{$99.38$}& \multirow{2}{*}{$19.603$} \\ 
  &  &  & $271.33$ &  & $128.11$ & & \\ 
 \hline\hline
 \multirow{4}{*}{CycleIK$_{GAN}$} & \multirow{2}{*}{Small} & \multirow{2}{*}{$2.892$} & $0.602$ / & \multirow{2}{*}{$0.266$} & $ 0.046$ / &\multirow{2}{*}{$92.07$}& \multirow{2}{*}{$0.458$} \\
  &  &  & $11.87$ &  & $2.82$ & &\\
 
  & \multirow{2}{*}{Full} & \multirow{2}{*}{$2.795$} & $0.7$ / & \multirow{2}{*}{$0.563$} & $ 0.134$/ & \multirow{2}{*}{$94.77$}&\multirow{2}{*}{$0.448$}\\
  &  &  & $10.84$ &  & $3.56$ & &\\
 \hline\hline
 \multirow{3}{*}{CycleIK$_{GAN}$} & \multirow{2}{*}{Small} & \multirow{2}{*}{$0.525$} & $6.84 \cdot 10^{-6}$ / & \multirow{2}{*}{$0.308$} & $4.22 \cdot 10^{-4}$ / & \multirow{2}{*}{$98.46$}& \multirow{2}{*}{$19.922$}\\ 
 \multirow{3}{*}{w. Gaikpy} &  &  & $169.33$ &  & $127.27$ &  &\\ 
 
 & \multirow{2}{*}{Full} & \multirow{2}{*}{$0.4$} & $9.34 \cdot 10^{-6}$ / & \multirow{2}{*}{$0.407$} & $7.58 \cdot 10^{-4}$ / & \multirow{2}{*}{$98.97$}&\multirow{2}{*}{$19.572$} \\ 
  &  &  & $366.22$ &  & $133.89$ & & \\ 
 \hline\hline

\multirow{4}{*}{Gaikpy} & \multirow{2}{*}{Small} & \multirow{2}{*}{$0.113$} & $5.16\cdot 10^{-6}$ / & \multirow{2}{*}{$8.066$} & $0.09$ / & \multirow{2}{*}{$93.33$}&\multirow{2}{*}{$1022.534$} \\ 
  &  &  & $62.83$ &  & $139.45$ & & \\ 

 & \multirow{2}{*}{Full} & \multirow{2}{*}{$0.062$} & $3.19 \cdot 10^{-6}$ / & \multirow{2}{*}{$5.969$} & $ 0.03$ / & \multirow{2}{*}{$96.06$}&\multirow{2}{*}{$1106.849$} \\ 
  &  &  & $100.83$ &  & $143.06$ & & \\ 
   \hline\hline

\multirow{4}{*}{BioIK} & \multirow{2}{*}{Small} & \multirow{2}{*}{$33.487$} & $1.24\cdot 10^{-6}$ / & \multirow{2}{*}{$7.625$} & $1.39\cdot 10^{-6}$ / & \multirow{2}{*}{$98.72$}& \multirow{2}{*}{$1$} \\ 
  &  &  & $654.83$ &  & $142.48$ & &  \\ 

 & \multirow{2}{*}{Full} & \multirow{2}{*}{$41.468$} & $9.93\cdot 10^{-6}$ / & \multirow{2}{*}{$8.349$} & $1.54\cdot 10^{-6}$ / & \multirow{2}{*}{$98.05$}& \multirow{2}{*}{$1$} \\ 
  &  &  & $575.57$ &  & $147.08$ & & \\ 

 \hline
\end{tabular}
}
\label{tb:final_results}
\vspace{-10pt}
\end{table}

For the GAN, 500 solutions for the same pose were generated, and the average error for every pose was calculated before the mean was taken over the whole test set. For all other methods, we only analyzed the error of the best solution for every test pose. For the GAN results, the average error of the best solution for every test pose is the average minimum error reported in Table~\ref{tb:final_results}.

%\vspace{-15pt}

It can be seen that the GAN model performs better for the Full workspace, while the MLP performs better for the Small workspace. The average error of the GANs is between three to ten times higher than for the MLPs. However, it was possible to improve the solutions of the GANs as well as the MLPs through optimization with Gaikpy. In general, the orientation errors of the MLPs increased while the positional errors decreased. Moreover, while the average maximum error of the GANs is near the upper limit we defined for the error, which is generally good as it indicates that most solutions are within the error limit, the success rate of the GANs can only compete with BioIK and the CycleIK MLP model through the genetic optimization. Both the MLP and GAN models can be deployed directly to real hardware without further optimization, as the positional error stays far below 1cm on average.

The standalone Gaikpy method shows a lower average positional error than BioIK and a similar to slightly lower rotational error. The divergent success definition of BioIK is the reason that its success rate of over $98\%$ outperforms the success rate of Gaikpy by about $3$-$5\%$, while the average error of BioIK is tremendously higher. When Gaikpy is seeded with the neural models from CycleIK, the error of the solutions can be reduced by around $60\%$ to $90\%$ while the timeout of the genetic algorithm can be reduced by over $98\%$, enabling the neuro-genetic method to directly compete with BioIK regarding success rate as well as average error. The standalone Gaikpy method overcomes both neuro-only architectures as well as the Gaikpy variant that was seeded with the GAN solutions with regard to the positional error. In contrast, Gaikpy's orientation error is higher than for all CycleIK setups, which indicates that the seeding with neural solutions increases Gaikpy's performance with regard to the orientation.

\vspace{-10pt}
\section{Conclusion}
\vspace{-7pt}
This work presented two novel neuro-inspired architectures for the inverse kinematics task that deliver state-of-the-art performance when compared to other bio-inspired methods. We showed that the neuro-only architectures are precise enough to be directly deployed to real-world robots. It was also shown that the solutions from the GAN, as well as the MLP architecture, can additionally be used as seeds for a genetic algorithm. The results showed that seeding the GA with the CycleIK output did not only improve the Cartesian precision of the neural solutions, but also reduced the runtime of the GA by over $98\%$. The weighted multi-objective function that was applied during the training of the MLP proved to successfully support the training and made it possible to influence the kinematic behavior of the model. Finally, the importance of the presented normalizing-flow method for the IK task is underlined, as the GAN model reaches a similar precision as IKFlow and therefore has better performance than most neuro-inspired IK approaches. CycleIK will be utilized for more sophisticated experimental setups in the future, such as collision-free motion planning in human-robot interaction and multi-modal grasping. 

\vspace{-10pt}
\subsubsection{Acknowledgements}
The authors gratefully acknowledge support from the
DFG (CML, MoReSpace, LeCAREbot), BMWK (SIDIMO, VERIKAS), and the European
Commission (TRAIL, TERAIS).  

\vspace{-10pt}
%--------------------------------------------------------------

%
% ---- Bibliography ----
%
% BibTeX users should specify bibliography style 'splncs04'.
% References will then be sorted and formatted in the correct style.
%
% \bibliographystyle{splncs04}
% \bibliography{mybibliography}
%
\def\url#1{}
\bibliographystyle{splncs04} % We choose the "plain" reference style
\bibliography{cycleik}
\end{document}